%% file: main.tex
\documentclass[10pt,twocolumn,letterpaper]{article}

\usepackage[pagenumbers]{iccv} 

\usepackage{pifont}
\usepackage{multirow} 
\usepackage{graphicx}
\usepackage{caption}
\definecolor{grey}{rgb}{0.5, 0.5, 0.5}  
\usepackage{array}
\usepackage{colortbl}
\usepackage{tabularray}


\definecolor{iccvblue}{rgb}{0.21,0.49,0.74}
\usepackage[pagebackref,breaklinks,colorlinks,allcolors=iccvblue]{hyperref}

\def\paperID{***} 
\def\confName{ICCV}
\def\confYear{2025}

\title{VLRMBench: A Comprehensive and Challenging Benchmark for Vision-Language Reward Models}

\author{
Jiacheng Ruan$^{1,}$\thanks{Jiacheng Ruan and Wenzhen Yuan are equal contributors. The work is in progress.} \quad Wenzhen Yuan$^{1,*}$ \quad Xian Gao$^{1}$ \quad Ye Guo$^{2}$ \\ Daoxin Zhang$^{2}$ \quad Zhe Xu$^{2}$ \quad Yao Hu$^{2}$ \quad Ting Liu$^{1}$ \quad Yuzhuo Fu$^{1,}$\thanks{Yuzhuo Fu is the corresponding author.}\\
$^{1}$Shanghai Jiao Tong University \\
$^{2}$Xiaohongshu Inc.\\
{\tt\small jackchenruan@sjtu.edu.cn}
}

\begin{document}
\maketitle
\input{sec/0_abstract}    
\input{sec/1_intro}

\input{sec/2_relatedwork}
\input{sec/3_benhmark}

\input{sec/4_exp}

\input{sec/5_analysis}
\input{sec/6_conclusion}

{
    \small
    \bibliographystyle{ieeenat_fullname}
    \bibliography{main}
}

\end{document}

%% file: sec/0_abstract.tex
\begin{abstract}
Although large visual-language models (LVLMs) have demonstrated strong performance in multimodal tasks, errors may occasionally arise due to biases during the reasoning process. 
Recently, reward models (RMs) have become increasingly pivotal in the reasoning process. Specifically, process RMs evaluate each reasoning step, outcome RMs focus on the assessment of reasoning results, and critique RMs perform error analysis on the entire reasoning process, followed by corrections. However, existing benchmarks for vision-language RMs (VLRMs) typically assess only a single aspect of their capabilities (e.g., distinguishing between two answers), thus limiting the all-round evaluation and restricting the development of RMs in the visual-language domain. To address this gap, we propose a comprehensive and challenging benchmark, dubbed as \textbf{VLRMBench}, encompassing 12,634 questions. VLRMBench is constructed based on three distinct types of datasets, covering mathematical reasoning, hallucination understanding, and multi-image understanding. We design 12 tasks across three major categories, focusing on evaluating VLRMs in the aspects of process understanding, outcome judgment, and critique generation. Extensive experiments are conducted on 21 open-source models and 5 advanced closed-source models, highlighting the challenges posed by VLRMBench. For instance, in the `Forecasting Future', a binary classification task, the advanced GPT-4o achieves only a 76.0\% accuracy. Additionally, we perform comprehensive analytical studies, offering valuable insights for the future development of VLRMs. We anticipate that VLRMBench will serve as a pivotal benchmark in advancing VLRMs. Code and datasets will be available at https://github.com/JCruan519/VLRMBench.

\end{abstract}

%% file: sec/1_intro.tex
\section{Introduction}
\label{sec:intro}

Recently, driven by the development of large language models (LLMs) \cite{llama2,llama3onlyL,qwen2,qwen25,llamamoe}, large visual-language models (LVLMs) have also achieved notable progress \cite{lvlmsurvey}. They have been broadly applied across domains such as medical imaging, remote sensing, autonomous driving, etc. \cite{LvlmMedicalSurvey,LvlmAutoDriveSurvey,skyeyegpt,camobj,ftiibench}. However, despite these remarkable achievements, they still encounter limitations in vision-language (VL) reasoning tasks, primarily due to insufficient reasoning depth and the absence of self-correction mechanisms \cite{mulberry,insightv}. To mitigate this issue, reward models (RMs) have been introduced to detect errors in model responses, thereby enhancing the performance of LVLMs \cite{criticv}.

During both the training and inference phases, RMs play a pivotal role. Firstly, prior to training, RMs can be employed to filter high-quality samples and facilitate the construction of automated data synthesis pipelines \cite{helpsteer2,DataSelectionSurvey,vlfeedback}. Secondly, during training, preference optimization techniques (e.g., RLAIF \cite{rlaif}) align the model's preferences with RMs, thereby enhancing the quality of the generated content. Finally, during inference, test-time scaling techniques, aided by RMs, are widely applied \cite{vlmcts,visvm}, trading additional inference time for improved model performance. In the above processes, RMs can be employed not only to evaluate and score the reasoning steps and final answers, but also to generate valuable feedback on the reasoning process. Therefore, it is essential to propose a benchmark for evaluating the various aspects of the performance of RMs. However, as shown in Table \ref{tab:comparison}, existing benchmarks predominantly focus on language domain, and benchmarks for VLRMs typically evaluate only a single capability (e.g., the quality comparison of multiple reasoning processes). These singular and inadequately challenging benchmarks fail to reveal the potential flaws of VLRMs, thereby limiting the development of RMs in the VL domain.

\begin{table}[!t]
\centering\footnotesize
\renewcommand{\arraystretch}{0.8}
\setlength{\tabcolsep}{3pt}
\begin{tabular}{lccc}
\toprule
&  \textbf{\# Tasks} & \textbf{Step Eval.?}    & \textbf{Test Size} \\
\midrule
MR-GSM8K \cite{MRGSM8k} & 1 &\ding{51}  & 2,999  \\
RMBench \cite{RMBench} & 1 &\ding{55} & 1,327 \\
CriticBench \cite{CriticBench} & 1 & \ding{55}  & -  \\
MathCheck-GSM \cite{MathCheck_GSM}   & 1 & \ding{51}   & 516 \\
MR-Ben \cite{mrben}  & 1 &\ding{51} & 5,975  \\
ProcessBench \cite{qwenprocessbench}  & 1 & \ding{51}  & 3,400  \\
PRMBench \cite{PRMBench} & 9 &\ding{51}  & 6,216  \\
VLRewardBench* \cite{VLRewardBench}  &1 &\ding{55}  &1,250 \\
MultimodalRewardBench* \cite{mmrewardbench}  &1 &\ding{55}  &5,211 \\
\midrule
\textsc{\textbf{VLRMBench}}*  &\textbf{12} &\ding{51}  &\textbf{12,634} \\
\bottomrule
\end{tabular}
\caption{Comparison between our VLRMBench and others. * represents the benchmark for vision-language domain. `Step Eval.?' refers to whether the step-level labels are included.}
\label{tab:comparison}
\end{table}

To bridge this gap, we propose VLRMBench, a benchmark that is both comprehensive and challenging, specifically designed for the VL domain. We introduce three distinct data categories: mathematical reasoning, hallucination understanding, and multi-image understanding, encompassing eight widely used datasets. Moreover, to obtain high-quality samples containing reasoning processes, we develop a collaborative data filtering and generation pipeline. Specifically, we first utilize a small LVLM to filter samples along two dimensions: quality and difficulty. Subsequently, QVQ-72B-preview \cite{qvq-72b-preview}, which excels in VL reasoning, is employed to generate reasoning processes. Next, we use the advanced GPT-4o \cite{gpt4} to partition the reasoning steps and perform an initial validation of correctness. Finally, we apply rule-based filtering and manual inspection to correct errors in the reasoning process, yielding 1,000 samples that serve as the foundation for task construction. Based on these samples, we introduce three themes: step-based, outcome-based, and criticism-based, encompassing 12 diverse tasks and 12,634 questions. These tasks can comprehensively evaluate VLRMs in the aspects of process understanding, outcome judgment, and criticism generation, thus laying the groundwork for the development of VLRMs.

We conducted extensive experiments on 21 open-source models (from 2B to 90B) and 5 closed-source models, such as GPT-4o and Claude-3.5-Sonnet \cite{Claude35Sonnet}, across 12 tasks. The results underscore the challenges of VLRMBench and demonstrate the narrowing gap between open-source and closed-source models. Specifically, in eight step-based tasks, even the advanced GPT-4o achieved only an average F1-Score of 62.4\%. In two outcome-based binary classification tasks, GPT-4o attained an average accuracy of only 66.3\%. Furthermore, in two criticism-based tasks, Qwen2.5-VL-72B \cite{Qwen2.5VL} outperformed GPT-4o, with an average win rate of 53.3\%, showcasing the superior performance of state-of-the-art (SOTA) open-source models. In addition, we performed extensive analytical studies, providing valuable insights into the development of VLRMs.

The main contributions can be summarized as follows: \textbf{1)} We introduce VLRMBench, the first benchmark that integrates both comprehensiveness and challenge for the evaluation of VLRMs. VLRMBench is constructed from 1,000 high-quality samples curated through our collaborative data filtering and generation pipeline, covering areas such as mathematical reasoning, hallucination understanding, and multi-image understanding. \textbf{2)} We designed 12 tasks, comprising a total of 12,634 carefully crafted questions, capable of evaluating models across three aspects: process understanding, outcome judgment, and critique generation. These diverse tasks enable targeted evaluations and reveal limitations of VLRMs. \textbf{3)} We conducted extensive experiments on VLRMBench using 21 open-source models and 5 closed-source models, highlighting the challenges presented by VLRMBench and offering comprehensive analysis and guidance for the development of VLRMs.

%% file: sec/2_relatedwork.tex
\section{Related Works}
\label{sec:relatedworks}

\subsection{Reward Models (RMs)}
Techniques such as Reinforcement Learning with Human Feedback (RLHF) utilize RMs that offer feedback for large models, aligning them with human preferences in terms of accuracy, utility, and safety \cite{rlhfRM,rlhfRM2}. RMs can generally be categorized into three types based on their objectives. Process-based RMs \cite{zhang2025lessons, ma2023let} evaluate intermediate reasoning steps to ensure logical consistency and the correctness of each step. Outcome-based RMs \cite{cai2024internlm2, liu2024skywork, lou2024uncertainty} assess the quality of the final output. In contrast, Critique-based RMs \cite{kim2023solar, wang2024self, chen2024self} emphasize offering textual feedback or critiques aimed at enhancing the model's performance. However, RMs are not always reliable \cite{PRMBench}. Therefore, it is imperative to establish a benchmark that facilitates a comprehensive evaluation of RMs.

\begin{figure*}[!t]
    \centering
    \includegraphics[width=0.99\linewidth]{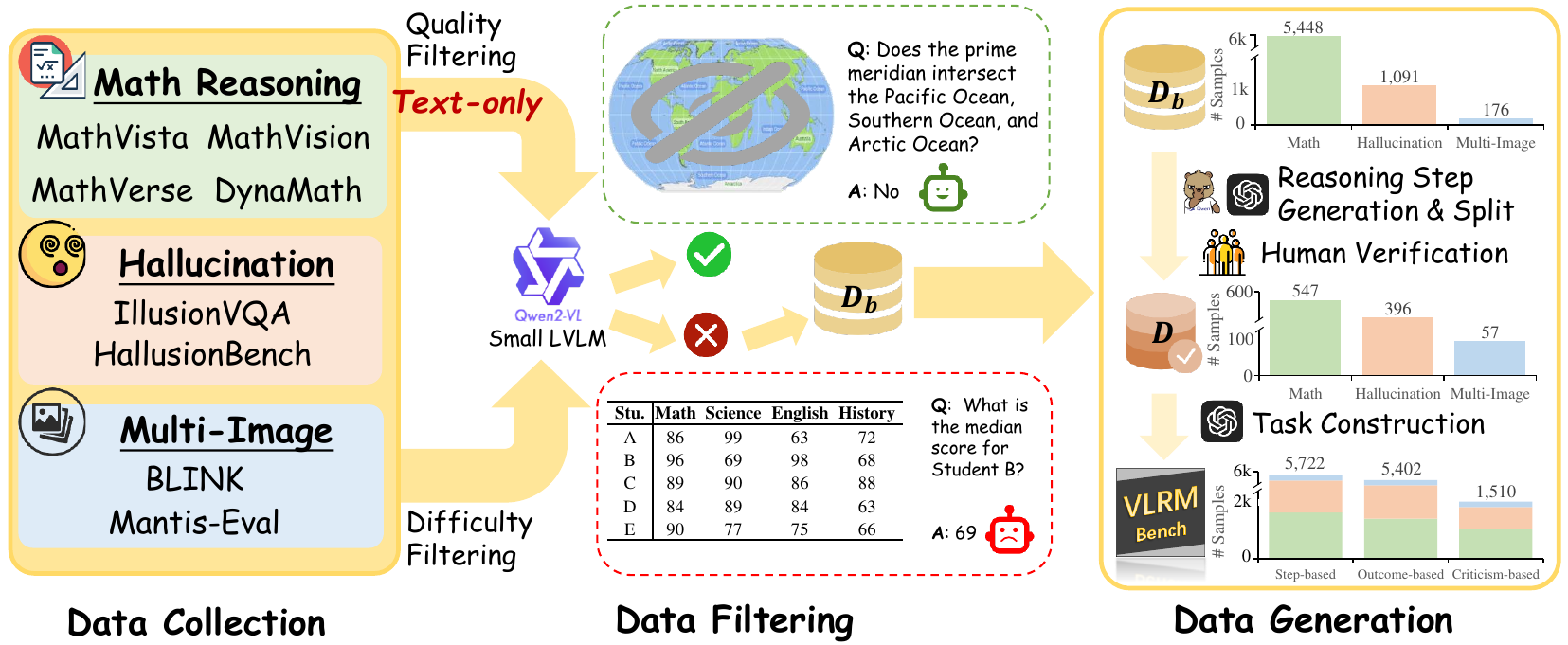}
    \caption{The collaborative data filtering and generation pipeline of VLRMBench. During the filtering stage, Qwen2VL-7B is tasked with answering questions under two conditions: without and with image input. Questions for which the model provides incorrect answers are retained. Subsequently, QVQ-72B-preview and GPT-4o are employed to generate reasoning processes and construct tasks for three themes.}
    \label{fig:main}
\end{figure*}

\subsection{Benchmarks of Reward Models}

In recent years, several benchmarks for evaluating the capabilities of RMs have been proposed, driving the development of RMs in the field of natural language. For example, MR-GSM8k \cite{MRGSM8k} redefines the model's role from a solver to a grader, necessitating the evaluation of the correctness of problem-solving steps. CriticBench \cite{CriticBench} incorporates the Generate-Critique-Revise framework to assess the reasoning capabilities of models. ProcessBench \cite{qwenprocessbench} emphasizes step-by-step error detection in mathematical derivations, thereby increasing the granularity of the evaluation. PRMBench \cite{PRMBench}, based on PRM800K \cite{prm800k}, employs GPT-4o for data optimization and constructs a fine-grained evaluation dataset comprising 6,216 instances. PRMBench covers nine common error types and provides a comprehensive assessment of Process RMs.

The aforementioned benchmarks predominantly focus on the language modality and do not adequately address the evaluation needs of LVLMs. To this end, \cite{VLRewardBench} introduces VLRewardBench, a benchmark for VLRMs containing 1,250 test samples. However, VLRewardBench remains limited to Pairwise Comparison, wherein only a judgment of superiority or inferiority is made between two responses, failing to incorporate more diverse and enriched task types. Consequently, we propose VLRMBench, which encompasses three core themes and twelve distinct tasks, providing a holistic evaluation of VLRMs' overall performance and establishing a foundation for their future advancement.

%% file: sec/3_benhmark.tex
\section{VLRMBench}
\label{sec:vlrmbench}

In this section, we initially present an overview of the pipeline for constructing VLRMBench, along with the data statistics information, followed by an in-depth description of the task design.

\subsection{Overall pipeline of VLRMBench}

In Figure \ref{fig:main}, we present a progressive and intuitive pipeline, which is designed to construct VLRMBench in three stages: data collection, filtering, and generation. During the collection phase, we integrated four mathematical, two hallucination, and two multi-image understanding datasets, totaling 16,550 samples and 19,343 images. In the filtering phase, to select high-quality and challenging samples, we introduced a joint filtering mechanism. Specifically, we initially employed Qwen2VL-7B\footnote{Qwen2VL-7B ranks in the top 25\% of the OpenCompass leaderboard \cite{vlmevalkit}. With moderate model capabilities and a small model size, it can efficiently filter out low-quality and low-difficulty samples.} \cite{Qwen2-VL} to answer questions without image information. If the model answered correctly, it indicated that the sample was of low quality \cite{MMStar}. Subsequently, we incorporated the image information into the question and instructed the model to answer it. If the model answered correctly, it suggested that the question was too simple to be a challenge. Consequently, we retained only those samples that could not be answered correctly in either question-answering mode, resulting in a total of 6,715 samples, which formed our basic database $D_b$.

\begin{figure}[!t]
    \centering
    \includegraphics[width=0.9\linewidth]{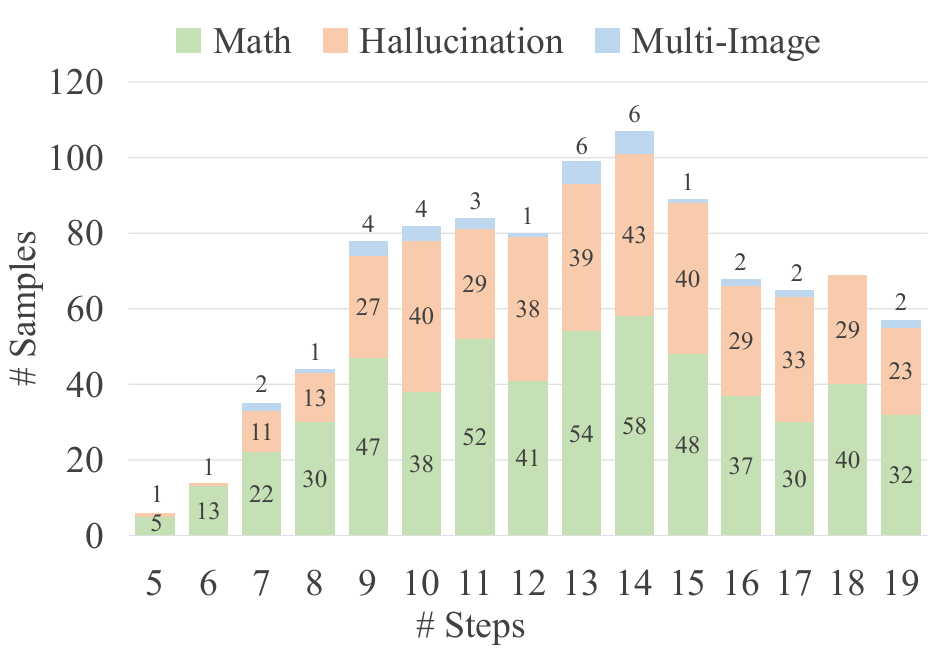}
    \caption{The distribution of the number of reasoning steps for different data sources.}
    \label{fig:bar_chart}
\end{figure}

In the generation phase, we utilized QVQ-72B-preview \cite{qvq-72b-preview} to generate reasoning process for the samples in $D_b$, discarding those where QVQ-72B-preview yielded incorrect answers, as verified via comparison with the standard answers. Subsequently, GPT-4o was employed to perform step-level segmentation of QVQ-72B-preview’s reasoning process, grouping semantically similar and logically related sentences into the one step. Following this, we obtained 3,335 samples with reasoning segmented at the step level. Furthermore, based on specific criteria\footnote{We excluded samples where the number of reasoning steps exceeded the average (i.e. 20) or was fewer than 5.}, we excluded samples with excessively long or short reasoning steps and ensured a balanced distribution across the three sample types. Finally, to ensure the quality of the samples, a manual review was conducted. 
Specifically, three PhD volunteers were recruited to verify the high-quality reasoning processes generated by QVQ-72B-preview. If any volunteer identified an error in the reasoning process, all three volunteers would collaboratively review the sample and correct any potential mistakes. 
Additionally, examples of manual corrections applied to the samples could be found in the Appendix. 
After completing the three aforementioned stages, 1,000 high-quality samples were retained for constructing the specific task, denoted as $D$. For more detailed statistical information, please refer to Table \ref{tab:samples_sources} and Figure \ref{fig:bar_chart}, which provide a description of the sample sources, the average word count, and the number of reasoning steps.

\begin{table}[!t]
    \centering
    \setlength{\tabcolsep}{0.5pt}
    \begin{tabular}{llcc}
    \toprule
    \textbf{Category} & \textbf{Dataset} & \textbf{\#Samples} & \textbf{\#Avg.Words} \\
    \midrule
    \multirow{4}{*}{Math} & MathVista \cite{lu2023mathvista} & 142 & 696 \\
                          & MathVision \cite{wang2025measuring} & 28 & 1,021 \\
                          & MathVerse \cite{zhang2024mathverse} & 126 & 742 \\
                          & DynaMath \cite{zou2024dynamath}   & 251 & 691 \\
    \midrule
    \multirow{2}{*}{Hallucination}  & IllusionVQA \cite{shahgir2024illusionvqa} & 62 & 736 \\
                                     & HallusionBench \cite{guan2024hallusionbench}  & 334 & 605 \\
    \midrule
    \multirow{2}{*}{Multi-Image}    & BLINK \cite{fu2024blink} & 23 & 790 \\
                                    & Mantis-Eval \cite{Jiang2024MANTISIM} & 34 & 592 \\
    \bottomrule
    \end{tabular}
    \caption{Statistics on the sample sources of VLRMBench. We also counted the average number of words in the reasoning process under different data sources.}
    \label{tab:samples_sources}
\end{table}

\subsection{Step-based tasks}

The step-based theme consists of 8 tasks designed to assess the reasoning process understanding ability of VLRMs. Under this theme, models will be provided with a modified reasoning process and are expected to output a sequence of 0s and 1s to indicate whether each reasoning step contains deviated content.

\begin{figure*}[!t]
    \centering
    \includegraphics[width=0.99\linewidth]{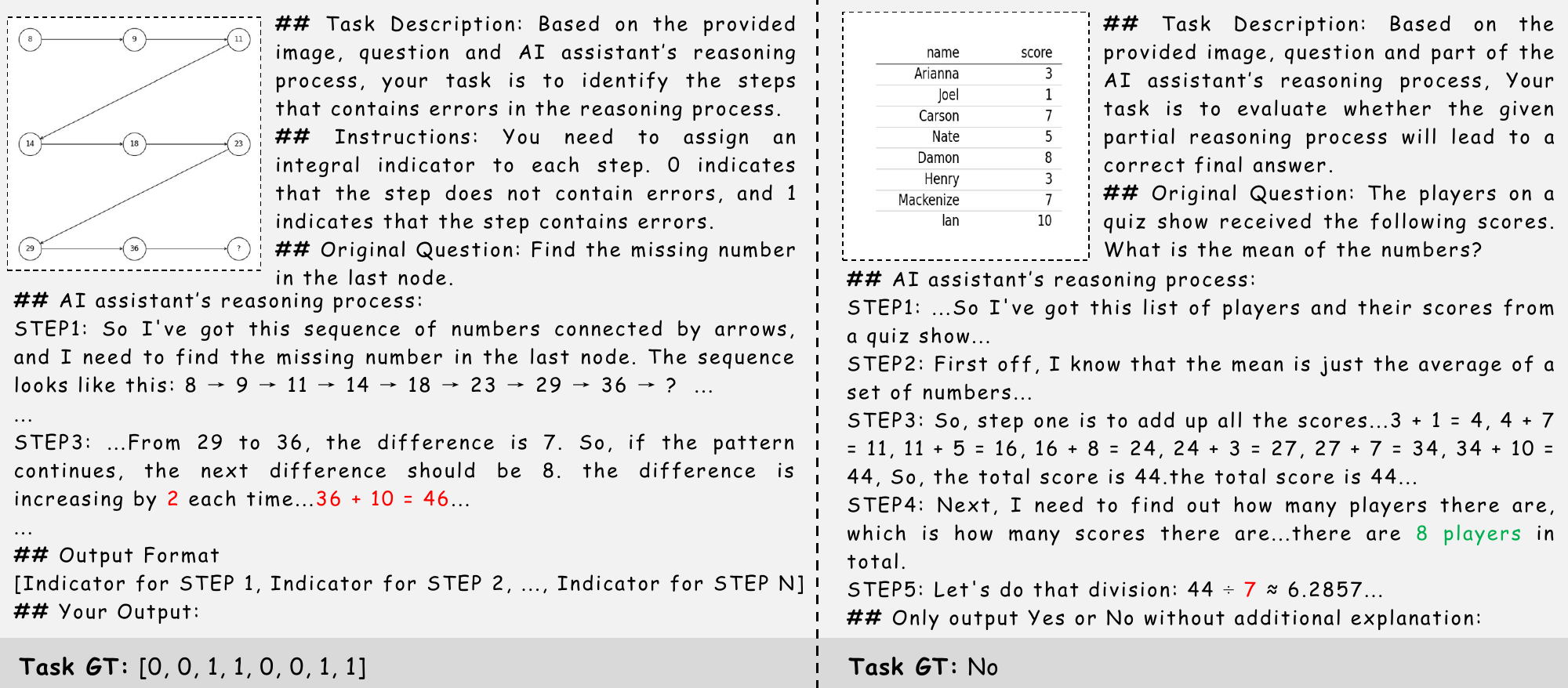}
    \caption{Specific input examples for VLRMBench. Left: An example of the SC task. VLRMs are required to understand each reasoning step and output a sequence, where `1' indicates the presence of erroneous information in the current reasoning step, and `0' indicates its absence. Right: An example of the FF task. VLRMs are tasked with assessing the correctness of the final result based on the reasoning process of the preceding steps.}
    \label{fig:maincase}
\end{figure*}

\paragraph{Step correctness (SC)} 
For a Vision-Language Reward Model (VLRM), detecting errors at each step of the reasoning process is a critical capability \cite{PRMBench, prm800k}. Building on this, we propose the SC task. Specifically, given the reasoning process $R_n$ of a sample from $D$, where $n$ denotes the total number of reasoning steps, we inject errors into $m$ of the $n$ steps ($0<m<n$) of $R_n$ using the advanced GPT-4o, to simulate potential errors during reasoning. As shown in Figure \ref{fig:maincase} Left, it is an example of the SC task.

\paragraph{Redundancy Detection (RD)} 
Existing reasoning models often produce redundant information when answering questions, leading to an increase in output token length, which is both inefficient and environmentally unfriendly. To this end, we propose the RD task, which requires VLRMs to identify redundant information generated during the reasoning process. Specifically, based on $D$, we introduce redundant descriptions in certain reasoning steps of each sample. For instance, we incorporate descriptions from $R_{i-j}$ into $R_{i}$ to intentionally create redundancy. Furthermore, we employ GPT-4 to generate paraphrased versions of the content from $R_{i-j}$ and incorporate them into $R_{i}$, thereby amplifying redundancy. Unlike the SC task, the introduction of redundant information does not compromise the correctness of the final answer.

\paragraph{Confidence Misdirection (CM)} 
When confident and assertive terms, such as `definitely' and `without a doubt', are present in an incorrect reasoning step, is it possible for VLRMs to be misled into perceiving the erroneous reasoning as correct? To investigate this issue, we propose the CM task, which is derived from the SC task. Specifically, for each sample in the SC task, we incorporate confidence-boosting terms into the erroneous reasoning step to mislead VLRMs into perceiving the step as correct, thereby evaluating the robustness of VLRMs.

\paragraph{Existential Hallucination (EH) and Attribute Hallucination (AH)}

When answering questions using LVLMs, hallucination phenomena often arise \cite{LvlmHallucinationSurvey}. Although other parts of the reasoning process may be correct, the introduction of hallucinatory errors or the omission of key entities can lead to biased final results. To evaluate whether VLRMs are capable of detecting hallucinations introduced during the reasoning process, we have designed two tasks on the top of $D$. The first task, EH, assesses the ability of the VLRM to identify entities that were incorrectly introduced during reasoning process but do not exist in the image, or to detect elements that may have been overlooked, such as objects, people, etc. The second task, AH, examines the VLRM's capability to accurately identify the attributes of entities, particularly in cases where LVLMs wrongly attribute properties to an entity that it does not possess or when attribute descriptions conflict with the visual content of the image. Such errors are typically reflected in misjudgments regarding object attributes, such as size, color, or shape. VLRMs must integrate image information effectively to accurately identify hallucination errors during the reasoning process.

\paragraph{Detail Error (DE)}

When addressing complex problems, the reasoning process often involves multiple details, particularly in the handling of numbers and symbols. Even if the overall reasoning appears to be sound, there may still be hidden errors in finer details (e.g., numerical errors in calculations or symbolic mistakes in formulas). These errors are often difficult to detect, yet they can affect the precision of the final results. The introduction of DE tasks serves to evaluate the sensitivity of VLRMs to reasoning details, as well as their ability to identify fine-grained errors.

\paragraph{Spatial Relationship (SR)}

During visual reasoning, LVLMs may incorrectly describe the spatial relationships between objects or entities, leading to erroneous results. For instance, LVLMs might misdescribe an object located at the top of the image as being at the bottom. Therefore, we select samples containing spatial relationship descriptions from $D$ to construct the SR task, aiming to evaluate VLRMs' understanding of spatial relationships in a image and their ability to detect spatial errors in a reasoning process.

\paragraph{Image Confusion (IC)}

In multi-image tasks, LVLMs are required to extract and associate relevant information from multiple images, which typically involves identifying various entities and their interrelations across different images. However, due to the presence of similar elements or backgrounds between images, LVLMs may erroneously associate an entity from one image with another image. This `image reference error' not only causes confusion but also potentially impacts the accuracy of subsequent reasoning. Thus, we filter samples related to multi-image understanding from $D$ to construct the IC task, which aims to evaluate whether VLRMs can identify image reference errors during the reasoning process.

\subsection{Outcome-based tasks}

The outcome-based theme encompasses two binary classification tasks, aiming to evaluate the ability of VLRMs to assess and predict the accuracy of outcomes.

\paragraph{Multi-solution Judgment (MJ)} 

During the training stage, preference optimization techniques, such as DPO \cite{dpo}, have become a widely used paradigm. These methods typically rely on the collection of preference samples. VLRMs are frequently employed to assess the quality of preference samples, thereby becoming a crucial component of preference optimization \cite{vlfeedback,rlaif}. To this end, we introduce the MJ task, which aims to investigate the discriminative ability of VLRMs regarding different reasoning processes for the same problem. This task is discussed as a core element in VLRewardBench \cite{VLRewardBench} and Multimodal RewardBench \cite{mmrewardbench}. Specifically, for the reasoning process $R$ of any question $Q$ in $D$, we introduce the reasoning process $R^\prime$ of the same question in the SC, RD and CM tasks, thereby constructing a triplet ($Q,R,R^\prime$). VLRMs are required to judge the quality of $R$ and $R^\prime$ and score them accordingly.

\paragraph{Forecasting Future (FF)}

When LVLMs perform inference, test-time scaling (TTS) can further improve their performance \cite{mulberry,vlmcts}. However, these techniques often introduce additional inference time, primarily due to the necessity of exploring multiple reasoning paths. If the correctness of the final answer could be predicted based on the reasoning process of the first m steps, $R_{[0:m]}$, this prediction could substantially expedite the TTS process. Therefore, the FF task is proposed, wherein VLRMs are required to assess the correctness of the final answer based on $R_{[0:m]}$. As shown in Figure \ref{fig:maincase} Right, this is an example of the FF task.

\subsection{Criticism-based tasks}
The criticism-based theme introduces two text generation tasks, in which VLRMs are required to analyze the potential causes of errors during the reasoning process and perform error correction.

\paragraph{Error Reason Analysis (ERA) and Error Correction (EC)}

Unlike Process RM and Outcome RM, which focus solely on evaluating processes or outcomes, Critique RM emphasizes the critique of the reasoning process, providing textual feedback that includes error analysis and correction. Consequently, we propose the ERA and EC tasks to assess the capabilities of VLRMs in these two aspects. In the ERA task, VLRMs are presented with erroneous reasoning steps derived from the SC task, and are tasked with identifying the error and analyzing its underlying cause. In contrast, the EC task does not require error cause analysis; rather, VLRMs directly correct the identified error and produce the corrected reasoning process.

%% file: sec/4_exp.tex
\begin{table*}[!t]
\centering\footnotesize
\setlength{\tabcolsep}{2.8pt}
\begin{tabular}{l|ccccccccc|ccc|ccc|ccc}
  \hline
\toprule
\multirow{2}{*}{\textbf{Model}} & \multicolumn{9}{c|}{\textbf{Step (5,722)}} & \multicolumn{3}{c|}{\textbf{Outcome (5,402)}} & \multicolumn{6}{c}{\textbf{Criticism (1,510)}} \\
  \cline{2-19}
 & \textbf{SC} & \textbf{RD} & \textbf{CM} & \textbf{EH} & \textbf{AH} & \textbf{DE} & \textbf{SR} & \textbf{IC} &  \textbf{W-Avg.} & \textbf{MJ} & \textbf{FF} & \textbf{W-Avg.} & \textbf{ERA\tiny{\textit{W}}} & \textbf{ERA\tiny{\textit{T}}} & \textbf{ERA\tiny{\textit{L}}} & \textbf{EC\tiny{\textit{W}}} & \textbf{EC\tiny{\textit{T}}} & \textbf{EC\tiny{\textit{L}}}\\
\hline
\# Sample & 998 & 997 & 994 & 946 & 944 & 659 & 126 & 58 & - & 2,986 & 2,416 & - &\multicolumn{3}{c|}{974}  &\multicolumn{3}{c}{536}  \\
\hline
\multicolumn{15}{l}{\textbf{\emph{Open-source LVLMs} $<$ 10B (Small group)}}\\
\hline
InternVL2.5-2B         & 33.8 & 34.1 & 30.0 & 22.4 & 23.8 & 33.5 & 19.9 & 30.1 & \cellcolor{grey!15}29.2 & 26.6 & 60.8 & \cellcolor{grey!15}41.9 & 4.8  & 6.7  & 88.5 & 2.3  & 10.0 & 87.6 \\
Qwen2VL-2B             & 28.3 & 26.2 & 20.9 & 17.8 & 19.1 & 26.6 & 23.8 & 26.4 & \cellcolor{grey!15}23.0 & 10.5 & 59.6 & \cellcolor{grey!15}32.5 & 0.2  & 0.1  & 99.7 & 2.5  & 5.8  & 91.7 \\
Qwen2.5VL-3B           & 29.9 & 32.7 & 23.2 & 27.2 & 26.9 & 45.5 & 25.1 & 42.6 & \cellcolor{grey!15}30.1 & 31.9 & 67.1 & \cellcolor{grey!15}47.6 & 7.1  & 12.1 & 80.8 & 4.8  & 21.2 & 73.9 \\
Phi-3.5-Vision (4.2B)  & 35.9 & 38.8 & 27.9 & 20.7 & 21.1 & 33.4 & 17.7 & 31.6 & \cellcolor{grey!15}29.2 & 12.1 & 65.2 & \cellcolor{grey!15}35.8 & 3.8  & 9.0  & 87.2 & 4.2  & 22.4 & 73.4 \\
Qwen2VL-7B             & 38.0 & 29.3 & 32.0 & 21.5 & 21.4 & 32.8 & 21.6 & 21.1 & \cellcolor{grey!15}28.8 & 52.0 & 64.7 & \cellcolor{grey!15}57.7 & 3.9  & 8.1  & 87.9 & 5.4  & 23.2 & 71.4 \\
Qwen2.5VL-7B           & 43.4 & 33.2 & 37.8 & 22.8 & 23.9 & 45.5 & 15.6 & 29.1 & \cellcolor{grey!15}33.4 & 26.0 & 70.7 & \cellcolor{grey!15}46.0 & 37.7 & 22.0 & 40.3 & 9.3  & 51.9 & 38.8 \\
Llava-OneVision-7B     & 31.8 & 25.7 & 21.5 & 16.0 & 17.2 & 24.8 & 16.1 & 20.7 & \cellcolor{grey!15}22.6 & 18.0 & 57.8 & \cellcolor{grey!15}35.8 & 3.9  & 10.4 & 85.7 & 7.1  & 34.4 & 58.5 \\
InternVL2.5-8B         & 36.6 & 28.4 & 31.1 & 21.9 & 21.2 & 36.5 & 15.1 & 24.1 & \cellcolor{grey!15}28.6 & 61.4 & 73.8 & \cellcolor{grey!15}66.9 & 34.5 & 21.9 & 43.6 & 6.2  & 36.1 & 57.7 \\
Ovis2-8B               & 47.7 & 28.2 & 37.8 & 25.2 & 25.7 & 39.8 & 17.0 & 25.8 & \cellcolor{grey!15}33.3 & 31.4 & 65.3 & \cellcolor{grey!15}46.6 & 23.2 & 25.9 & 50.9 & 14.3 & 63.5 & 22.2 \\
MiniCPM-V-2.6 (8B)     & 44.9 & 38.3 & 36.9 & 26.1 & 26.5 & 36.8 & 24.2 & 30.6 & \cellcolor{grey!15}34.6 & 37.9 & 62.5 & \cellcolor{grey!15}48.9 & 5.4  & 11.7 & 82.9 & 3.4  & 24.7 & 71.9 \\
MiniCPM-o-2.6 (8B)     & 41.0 & 35.3 & 37.3 & 29.6 & 29.0 & 36.7 & 28.3 & 30.1 & \cellcolor{grey!15}34.6 & 15.0 & 64.3 & \cellcolor{grey!15}37.0 & 9.2  & 19.8 & 70.9 & 7.6  & 33.9 & 58.4 \\
\rowcolor{grey!36} \textbf{Avg.}          & 37.4 & 31.8 & 30.6 & 22.8 & 23.3 & 35.6 & 20.4 & 28.4 & 29.8 & 29.3 & 64.7 & 45.2 & 12.2  & 13.4 & 74.4 & 6.1  & 29.7 & 64.1 \\

\midrule
\multicolumn{19}{l}{\textbf{10B $<$ \emph{Open-source LVLMs} $<$ 40B (Middle group)}}\\
\hline
Llama3.2-11B           & 33.9 & 27.1 & 29.7 & 20.2 & 20.0 & 31.9 & 17.1 & 25.8 & \cellcolor{grey!15}26.7 & 23.9 & 61.1 & \cellcolor{grey!15}40.5 & 4.2  & 10.3 & 85.5 & 5.4  & 34.7 & 59.8 \\
Ovis2-16B              & 48.5 & 47.9 & 47.9 & 40.6 & 35.8 & 44.5 & 21.9 & 28.4 & \cellcolor{grey!15}43.6 & 49.7 & 69.2 & \cellcolor{grey!15}58.4 & 30.9 & 26.6 & 42.5 & 10.4 & 69.3 & 20.3 \\
Ovis2-34B              & 65.3 & 51.1 & 64.5 & 54.5 & 51.6 & 59.6 & 30.4 & 49.0 & \cellcolor{grey!15}57.0 & 46.3 & 75.7 & \cellcolor{grey!15}59.4 & 45.9 & 26.6 & 27.5 & 15.4 & 76.4 & 8.1 \\
InternVL2.5-38B        & 61.0 & 49.6 & 62.6 & 55.0 & 51.7 & 61.4 & 34.6 & 48.1 & \cellcolor{grey!15}56.1 & 45.1 & 79.3 & \cellcolor{grey!15}60.4 & 21.1 & 24.5 & 54.4 & 12.5 & 56.6 & 30.9 \\
\rowcolor{grey!36} \textbf{Avg.}          & 52.2 & 43.9 & 51.2 & 42.6 & 39.8 & 49.4 & 26.0 & 37.8 & 45.9 & 41.3 & 71.3 & 54.7 & 25.5 & 22.0 & 52.5 & 10.9 & 59.3 & 29.8 \\

\midrule
\multicolumn{15}{l}{\textbf{\emph{Open-source LVLMs} $>$ 40B (Large group)}}\\
\hline
Llava-OneVision-72B    & 49.4 & 30.6 & 50.2 & 37.9 & 34.1 & 45.1 & 13.9 & 26.5 & \cellcolor{grey!15}40.4 & 32.2 & 69.7 & \cellcolor{grey!15}49.0 & 15.3 & 22.1 & 62.6 & 9.4  & 51.0 & 39.6 \\
Qwen2VL-72B            & 55.5 & 40.2 & 56.1 & 48.9 & 41.6 & 56.4 & 22.4 & 43.6 & \cellcolor{grey!15}48.8 & 37.7 & 72.1 & \cellcolor{grey!15}53.1 & 22.4 & 20.4 & 57.1 & 7.7  & 57.1 & 35.1 \\
Qwen2.5VL-72B          & 72.8 & 41.7 & 70.4 & 64.6 & 59.9 & 72.4 & 37.9 & 63.3 & \cellcolor{grey!15}62.6 & 65.6 & 80.2 & \cellcolor{grey!15}72.1 & 74.1 & 15.1 & 10.8 & 15.6 & 77.0 & 7.3 \\
QVQ-72B-preview        & 42.4 & 35.3 & 36.3 & 29.4 & 27.1 & 37.8 & 20.8 & 26.1 & \cellcolor{grey!15}34.3 & 49.2 & 38.4 & \cellcolor{grey!15}44.4 & 4.2  & 10.0 & 85.8 & 9.4  & 51.0 & 39.6 \\
InternVL2.5-78B        & 62.1 & 42.0 & 62.8 & 53.3 & 48.7 & 60.4 & 28.5 & 47.2 & \cellcolor{grey!15}54.0 & 48.6 & 78.8 & \cellcolor{grey!15}62.1 & 54.2 & 24.9 & 20.9 & 10.8 & 63.9 & 25.3 \\
Llama3.2-90B           & 40.8 & 38.8 & 41.3 & 34.2 & 29.4 & 35.4 & 17.0 & 21.1 & \cellcolor{grey!15}36.2 & 52.7 & 69.0 & \cellcolor{grey!15}60.0 & 9.9  & 18.6 & 71.4 & 9.5  & 56.8 & 33.7 \\
\rowcolor{grey!36} \textbf{Avg.}          &53.8 	&38.1 	&52.9 	&44.7 	&40.1 	&51.3 	&23.4 	&38.0 	&46.1 	&47.7 	&68.0 	&56.8 	&30.0 	&18.5 	&51.4 	&10.4 	&59.5 	&30.1   \\

\midrule
\hline
\multicolumn{15}{l}{\textbf{\emph{Closed-source LVLMs}}}\\
\hline
Claude-3.5-Sonnet      & 70.8 & 53.7 & 65.7 & 63.9 & 62.8 & 63.4 & 40.8 & 65.7 & \cellcolor{grey!15}62.9 & 82.2 & 75.1 & \cellcolor{grey!15}79.0 & 60.6 & 25.5 & 13.9 & 21.2 & 53.9 & 24.9 \\
GPT-4o                 & 73.7 & 50.6 & 66.6 & 57.6 & 58.6 & 71.8 & 48.2 & 64.6 & \cellcolor{grey!15}62.4 & 58.4 & 76.0 & \cellcolor{grey!15}66.3 & 0.0  & 100.0  & 0.0  & 0.0  & 100.0  & 0.0 \\
GLM-4V-Flash           & 37.2 & 33.5 & 32.9 & 28.2 & 27.7 & 41.8 & 22.9 & 20.2 & \cellcolor{grey!15}32.8 & 37.2 & 66.7 & \cellcolor{grey!15}50.4 & 34.2 & 24.3 & 41.5 & 6.9  & 54.1 & 39.0 \\
Gemini-2.0-Flash       & 69.3 & 55.9 & 66.1 & 65.6 & 62.3 & 63.4 & 46.8 & 67.8 & \cellcolor{grey!15}63.4 & 54.2 & 77.1 & \cellcolor{grey!15}64.4 & 32.8 & 22.5 & 44.7 & 7.1  & 52.5 & 40.3 \\
Gemini-2.0-FTE.-1219   & 67.9 & 55.5 & 65.7 & 65.9 & 62.3 & 63.2 & 46.4 & 67.1 & \cellcolor{grey!15}63.1 & 53.1 & 77.0 & \cellcolor{grey!15}63.8 & 37.4 & 22.6 & 40.0 & 6.8  & 50.2 & 43.1 \\
\rowcolor{grey!36} \textbf{Avg.}          & 63.8 	& 49.8 	& 59.4 	& 56.2 	& 54.7 	& 60.7 	& 41.0 	& 57.1 	& 57.0 	& 57.0 	& 74.4 	& 64.8 	& 33.0 	& 39.0 	& 28.0 	& 8.4 	& 62.1 	& 29.5  \\

\bottomrule
\hline
\end{tabular}
\caption{Results on the VLRMBench. For Step-based tasks, the F1-Score is the metric. For Outcome-based tasks, Accuracy is utilized as the metric. In the case of Criticism-based tasks, the metric is Win Rate, expressed as Win:Tie:Lose (Simplified as \textit{W}, \textit{T} and \textit{L}). Note that W-Avg. represents the weighted average. Evaluation prompt templates and some case studies could be found in the Appendix B and C.}
\label{tab:mainres}
\end{table*}

\section{Experiments}
\label{sec:exp}

\subsection{Models}

We conducted extensive experiments on 21 open-source models and 5 closed-source models. Specifically, the open-source models, which range from 2B to 90B, including Llama3.2 (11B/90B) \cite{llama3}, Llava-OneVision (7B/72B) \cite{Llava-OneVision}, Qwen2VL (2B/7B/72B) \cite{Qwen2-VL}, Qwen2.5VL (3B/7B/72B) \cite{Qwen2.5VL}, QVQ-72B-preview \cite{qvq-72b-preview}, Ovis2 (8B/16B/34B) \cite{Ovis2}, InternVL2.5 (2B/8B/38B/78B) \cite{InternVL2.5}, Phi-3.5-Vision (4.2B) \cite{Phi-3.5-Vision}, MiniCPM-o-2.6 (8B), and MiniCPM-V-2.6 (8B) \cite{MiniCPM}. For the closed-source models, we employed GLM-4V-Flash \footnote{https://open.bigmodel.cn/dev/api/normal-model/glm-4v. To our best knowledge, GLM-4V-Flash is the first free-to-use multimodal API model.}, GPT-4o \cite{gpt4}, Gemini-2.0-Flash/-Flash-thinking-exp-1219 \cite{gemini}, and Claude-3.5-Sonnet \cite{Claude35Sonnet}.

\subsection{Evaluation details}

We follow the `LVLM-as-a-Judge' paradigm and introduce different question-answering templates to assess the models \cite{VLRewardBench,mmrewardbench}. Specifically, for the SC, RD, CM, EH, AH, DE, SR, and IC tasks, the model takes a reasoning process as input and outputs a judgment regarding whether each reasoning step contains redundancies or other errors. For these eight tasks, we employ the weighted F1-Score as the evaluation metric, as defined in Equation \ref{eq:f1_w_score}, which better addresses class imbalance issues and ensures the model's performance on the minority class is adequately considered. For the MJ task, the model is provided with (good and bad) reasoning process pairs for the same question and outputs scores for both. If the model assigns a higher score to the good sample, it is deemed correct; otherwise, it is considered incorrect. For the FF task, the model takes several preceding steps of the reasoning process as input and outputs a judgment on whether a correct answer can ultimately be derived. Consequently, for the MJ and FF tasks, Accuracy is employed as the evaluation metric. For the ERA and EC tasks, the model’s output is a description of the error reasons in the reasoning process and the corrected reasoning process. We utilize the output of GPT-4o as the baseline and adopt Win Rate \cite{wr1alpacaeval2, wr2mtbench} as the evaluation metric. 

We configured the max new tokens to 4,096 across all tasks for the five closed-source models and QVQ-72B-preview. In contrast, for the other models, we set the max new tokens to 4,096 specifically for the ERA and EC tasks, while fixing it at 256 for the remaining tasks. To ensure the reproducibility of the experimental results, we set the temperature coefficient to 0 for all models. All experiments were conducted on 8 NVIDIA H800 GPUs.

\begin{equation}
    F1 = \frac{N_{\text{neg}}}{N_{\text{pos}} + N_{\text{neg}}} \cdot F1_{\text{pos}} + \frac{N_{\text{pos}}}{N_{\text{pos}} + N_{\text{neg}}} \cdot F1_{\text{neg}}
\label{eq:f1_w_score}
\end{equation}
where $N_{\text{pos}}$ and $N_{\text{neg}}$ represent the number of correct and incorrect steps in the reasoning process, respectively.

\subsection{Main Results}

The experimental results on VLRMBench are shown in Table \ref{tab:mainres}, leading to the following conclusions: 

\noindent\textbf{1) Among open-source models, surprisingly, the performance of Qwen2.5VL-72B is comparable to that of advanced closed-source models.} For instance, across the eight Step-based tasks, Qwen2.5VL-72B achieved an average F1-Score of 62.6\%, surpassing GPT-4o by 0.2\% and only slightly trailing Claude-3.5-Sonnet by 0.3\%. In the MJ and FF tasks, Qwen2.5VL-72B also outperformed GPT-4o by 7.2\% and 4.2\%, respectively. Notably, in the ERA task, Qwen2.5VL-72B attained a win rate of 74.1\% over GPT-4o, significantly outperforming Claude-3.5-Sonnet at 60.6\%. 

\noindent\textbf{2) LVLMs that have not undergone specific training show diminishing performance gains as the model size increases.} Specifically, for LVLMs in the Middle group, compared to those in the Small group, the performance gain in Step-based and Outcome-based tasks was 54.0\% and 21.0\%, respectively. In contrast, the corresponding performance gains in the Large group relative to the Middle group were only 0.4\% and 3.8\%. Even excluding QVQ-72B-preview from the Large group, the relative performance improvement remains limited, at 5.4\% and 8.4\%. This suggests that blindly increasing the size of LVLMs does not lead to significant performance improvements. Consequently, we recommend that future research prioritize training VLRMs at the Middle group scale initially.

\begin{table}[!t]
\setlength{\tabcolsep}{3pt}
\renewcommand{\arraystretch}{0.9} 
\centering
\begin{tabular}{l|ccc}
\toprule
\textbf{Order} & \textbf{Model} & \textbf{Avg. Score} & \textbf{Acc} \\
\midrule
\multirow{4}{*}{\textbf{Better first}} 
& Qwen2.5VL-3B & [9.0, 8.3] & 23.7 \\
& Llama3.2-11B & [8.6, 4.8] & 52.8 \\
& InternVL2.5-78B & [8.4, 7.4] & 38.6 \\
& GPT-4o & [9.8, 7.8] & 52.2 \\
\midrule
\multirow{4}{*}{\textbf{Better second}} 
& Qwen2.5VL-3B & [8.7, 9.0] & 21.4 \\
& Llama3.2-11B & [8.3, 5.6] & 8.4 \\
& InternVL2.5-78B & [8.1, 9.1] & 45.6 \\
& GPT-4o & [7.4, 8.1] & 43.1 \\
\bottomrule
\end{tabular}
\caption{The results on the MJ task with different sample order.}
\label{abl:mj}
\end{table}

\noindent\textbf{3) The Outcome-based tasks, being simple binary classification problems, remain challenging for most open-source models, and even closed-source models perform only marginally better than random guessing.} For open-source models across three size groups, the average accuracy for Outcome-based tasks is around 50\%, similar to random guessing. For advanced closed-source models, the average accuracy only exceeds random guessing by 14.8\%. 

\noindent\textbf{4) Deterministic vocabulary and hallucinated information can effectively mislead the VLRMs, reducing its ability to understand reasoning processes.} The CM task, based on the SC task, introduces deterministic vocabulary (e.g., must, definitely) into sentences with erroneous reasoning steps, while EH and AH tasks introduce hallucinated information during the reasoning process. Existing models, particularly in the Small group, demonstrate a degradation in F1-Score of 6.8\%, 14.6\%, and 14.1\% for CM, EH, and AH tasks, respectively, when compared to the SC task. This performance degradation remains significant even as model size increases. 

\noindent\textbf{5) Existing models show strong robustness to numerical errors but limited resilience to perturbations in spatial relationships and image references.} We observed that for Large group, the F1-Score in the DE task only decreased by 2.5\% compared to SC. However, for the SR and IC tasks, the performance dropped by 30.4\% and 15.8\%, respectively. Overall, our VLRMBench provide unique challenges that surpass traditional benchmarks, with even the most advanced models achieving only mediocre performance.

%% file: sec/5_analysis.tex
\section{Further Analysis}
\label{sec:ana}

\subsection{Is the o1-like model suitable to be a VLRM?}

QVQ-72B-preview \cite{qvq-72b-preview}, an open-source visual-language model, exhibits strong reasoning capabilities. However, it exhibits a potential drawback regarding the degradation of instruction-following ability. For instance, in the SC task, by incorporating instruction prompts, the sample's original question, and the reasoning process as input, we expect the VLRM to evaluate each reasoning step. However, QVQ-72B-preview often disregards the given instructions and directly answers the original question, leading to suboptimal performance on VLRMBench, with additional cases provided in the Appendix C. Therefore, as an o1-like model, it must not only demonstrate strong reasoning abilities but also exhibit robust instruction-following capabilities to be considered a viable candidate for a VLRM.

\subsection{Impact of sample order}

In the MJ task, VLRM scores two reasoning processes for the same question. To further investigate the positional bias, we employed two different sample orders, as shown in Table \ref{abl:mj}. Specifically, `Better first' places the superior reasoning process in the first position, while `Better second' places it in the second position. The results indicate that Llama3.2-11B consistently assigns a higher score to the reasoning process in the first position, irrespective of its quality. In contrast, the other three models exhibit the ability to adjust their scores based on the quality of the reasoning. 
This positional bias phenomenon is also observed in \cite{sampleorder}. As a result, in Table \ref{tab:mainres}, we utilized the sum of the two scores for the MJ task evaluation to ensure fairness.

\subsection{Performance on various reasoning length}

\begin{table}[!t]
\setlength{\tabcolsep}{3pt}
\renewcommand{\arraystretch}{0.9} 
\centering
\begin{tabular}{l|ccc}
\toprule
\textbf{Model} & \textbf{$\leq$10 steps} & \textbf{10-15 steps} & \textbf{$>$15 steps} \\
\midrule
Qwen2.5VL-3B & 35.6 & 30.1 & 25.6 \\
Llama3.2-11B & 42.9 & 32.8 & 28.4 \\
InternVL2.5-78B & 68.2 & 64.1 & 54.1 \\
GPT-4o & 77.8 & 75.2 & 68.6 \\
\bottomrule
\end{tabular}
\caption{Results on the SC task with different reasoning steps.}
\label{abl:length}
\end{table}

We further investigate the impact of the number of reasoning steps on the VLRMs' process understanding. As shown in Table \ref{abl:length}, we classified the number of reasoning steps in the SC task into three categories: the short group ($\leq$ 10 steps), with 263 samples; the middle group (10-15 steps), with 468 samples; and the long group ($>$ 15 steps), with 267 samples. Specifically, for all four models, a decrease in F1-score was observed as the number of reasoning steps increased. Notably, for the Llama3.2-11B model, the difference between the short and long groups reached 14.5\%, and even for the advanced GPT-4o, the F1-Score dropped from 77.8\% to 68.6\%. This suggests that current models are less capable of handling long information, underscoring the necessity of enhancing their ability to tackle long reasoning process when training VLRMs in the future.

\subsection{Impacts of In-Context Learning (ICL) setting}

In main experiments, we employ zero-shot setting by default. Furthermore, using the SC task as an example, we explore the effect of ICL on VLRMs. As presented in Table \ref{tab:fewshot}, the results indicate that providing contextual examples is more beneficial for small models but harms the performance of large models. Specifically, with the help of one-shot learning, Qwen2.5VL-3B's performance improved from 29.9\% to 31.5\%, and Llama3.2-11B saw an improvement of 8.3\%. However, for GPT-4o, one-shot learning led to a 2.0\% performance degradation, and InternVL2.5-78B’s performance decreased by 6.2\%. One possible explanation for this is that small models exhibit relatively weaker instruction-following capabilities, and ICL examples may help guide them in responding to queries in the designated format. In contrast, large models, which already possess strong comprehension abilities and adaptability to complex tasks, may find that providing contextual examples interferes with their inherent understanding process, resulting in performance decline. Thus, in the SC task, smaller models exhibit a greater reliance on ICL examples, whereas larger models tend to rely more on their pre-trained knowledge.

\begin{table}
\centering
\renewcommand{\arraystretch}{0.9} 
    \begin{tabular}{l|ccc}
    \toprule
    \textbf{Model} & \textbf{0-shot} & \textbf{1-shot} & \textbf{2-shot} \\
    \midrule
    Qwen2.5VL-3B & 29.9 & 31.5 & 31.1 \\
    Llama3.2-11B & 33.9 & 42.2 & 42.3 \\
    InternVL2.5-78B & 62.1 & 55.9 & 55.7 \\
    GPT-4o & 73.7 & 71.7 & 70.5 \\
    \bottomrule
    \end{tabular}
\caption{Results on SC task with various ICL few-shot numbers.}
\label{tab:fewshot}
\end{table}

\subsection{Impacts of test-time scaling}

\begin{figure}
    \centering
    \includegraphics[width=0.9\linewidth]{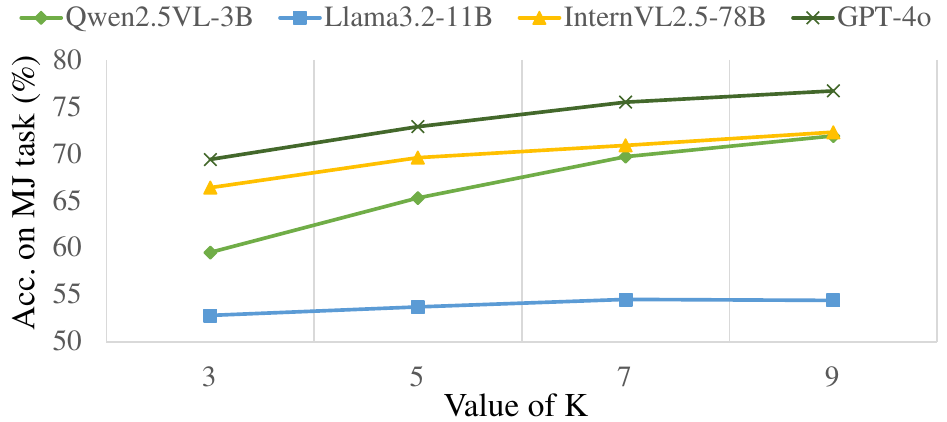}
    \caption{Performance changes with the test-time scaling.}
    \label{fig:voting}
\end{figure}

We investigated the impact of test-time scaling (i.e. voting method) for VLRMs on the MJ task. To ensure the diversity of the generated results, we set the temperature coefficient to 0.9 and performed K sampling. As illustrated in Figure \ref{fig:voting}, surprisingly, as the value of K increases, the accuracy of Qwen2.5VL-3B improved by 12.4\%. Notably, at K = 9, its performance was only 4.8\% lower than that of GPT-4o. These findings highlight the substantial effectiveness of test-time scaling techniques in enhancing VLRMs.

\subsection{Improve LVLMs with a VLRM}

\begin{table}[!t]
\renewcommand{\arraystretch}{0.85} 
\setlength{\tabcolsep}{1pt}
\centering
\begin{tabular}{cc|cccc}
\toprule
\textbf{LVLM} &\textbf{VLRM} & \textbf{MathVista} & \textbf{MathVerse} & \textbf{MathVision} & \textbf{Avg.} \\
\midrule
7B &7B   &10.4 &5.8  &4.7 &7.0\\
72B &7B  &11.8 &4.2  &3.2 &6.4\\
7B &72B  &21.8 &19.7  &21.5 &21.0\\
72B &72B &11.7 &9.3  &11.4 &10.8\\
\bottomrule
\end{tabular}
\caption{Results of the correction rate on math benchmarks.}
\label{tab:correctionrate}
\end{table}

To demonstrate the significant role of VLRMs in the reasoning process, as presented in Table \ref{tab:correctionrate}, we conducted experiments on three mathematical benchmarks using Qwen2.5VL-7B/72B. For example, in the second row of Table \ref{tab:correctionrate}, Qwen2.5VL-72B was employed as the LVLM to answer the mathematical questions, whereas Qwen2.5VL-7B was used as the VLRM to perform error reason analysis of the LVLM’s reasoning process. The analysis was subsequently fed back into the LVLM as feedback, allowing it to generate a second round of answers. Besides, we defined the correction rate, which is the ratio of the number of questions correctly answered in the second round after feedback to the number of questions incorrectly answered in the first round. The experimental results demonstrate that VLRM can provide valuable information to the LVLM through error analysis, thereby facilitating mistake correction, with an average gain of 11.3\%. These findings suggest that even LVLMs not trained on proprietary data can act as VLRMs, providing more correction possibilities during testing based on this intuitive feedback mechanism. This further underscores the potential value of developing efficient VLRMs.

%% file: sec/6_conclusion.tex
\section{Conclusions}

In this paper, we propose a comprehensive and challenging benchmark, named VLRMBench. This benchmark covers 3 themes and 12 tasks, comprising a total of 12,634 questions. Following the `LVLM-as-a-Judge' paradigm, we conduct systematic experiments and analyses on VLRMBench using 26 models, deriving several key insights: enhancing the performance of VLRMs in long-text comprehension, exploring various feedback patterns from VLRMs to LVLMs, and developing the application potential of proprietary VLRMs, etc. We hope that VLRMBench will serve as a standardized evaluation platform to propel the development of more efficient VLRMs. In future work, we plan to incorporate larger datasets specifically designed to train VLRMs.

%% file: main.bbl
\begin{thebibliography}{67}
\providecommand{\natexlab}[1]{#1}
\providecommand{\url}[1]{\texttt{#1}}
\expandafter\ifx\csname urlstyle\endcsname\relax
  \providecommand{\doi}[1]{doi: #1}\else
  \providecommand{\doi}{doi: \begingroup \urlstyle{rm}\Url}\fi

\bibitem[Abdin et~al.(2024)Abdin, Aneja, Awadalla, Awadallah, Awan, Bach, Bahree, Bakhtiari, Bao, Behl, et~al.]{Phi-3.5-Vision}
Marah Abdin, Jyoti Aneja, Hany Awadalla, Ahmed Awadallah, Ammar~Ahmad Awan, Nguyen Bach, Amit Bahree, Arash Bakhtiari, Jianmin Bao, Harkirat Behl, et~al.
\newblock Phi-3 technical report: A highly capable language model locally on your phone.
\newblock \emph{arXiv preprint arXiv:2404.14219}, 2024.

\bibitem[Achiam et~al.(2023)Achiam, Adler, Agarwal, Ahmad, Akkaya, Aleman, Almeida, Altenschmidt, Altman, Anadkat, et~al.]{gpt4}
Josh Achiam, Steven Adler, Sandhini Agarwal, Lama Ahmad, Ilge Akkaya, Florencia~Leoni Aleman, Diogo Almeida, Janko Altenschmidt, Sam Altman, Shyamal Anadkat, et~al.
\newblock Gpt-4 technical report.
\newblock \emph{arXiv preprint arXiv:2303.08774}, 2023.

\bibitem[Anthropic(2025)]{Claude35Sonnet}
Anthropic.
\newblock Claude 3.5: Sonnet, 2025.
\newblock Accessed: 2025-02-25.

\bibitem[Bai et~al.(2025)Bai, Chen, Liu, Wang, Ge, Song, Dang, Wang, Wang, Tang, et~al.]{Qwen2.5VL}
Shuai Bai, Keqin Chen, Xuejing Liu, Jialin Wang, Wenbin Ge, Sibo Song, Kai Dang, Peng Wang, Shijie Wang, Jun Tang, et~al.
\newblock Qwen2. 5-vl technical report.
\newblock \emph{arXiv preprint arXiv:2502.13923}, 2025.

\bibitem[Bai et~al.(2022)Bai, Kadavath, Kundu, Askell, Kernion, Jones, Chen, Goldie, Mirhoseini, McKinnon, et~al.]{rlhfRM}
Yuntao Bai, Saurav Kadavath, Sandipan Kundu, Amanda Askell, Jackson Kernion, Andy Jones, Anna Chen, Anna Goldie, Azalia Mirhoseini, Cameron McKinnon, et~al.
\newblock Constitutional ai: Harmlessness from ai feedback.
\newblock \emph{arXiv preprint arXiv:2212.08073}, 2022.

\bibitem[Cai et~al.(2024)Cai, Cao, Chen, Chen, Chen, Chen, Chen, Chen, Chen, Chu, et~al.]{cai2024internlm2}
Zheng Cai, Maosong Cao, Haojiong Chen, Kai Chen, Keyu Chen, Xin Chen, Xun Chen, Zehui Chen, Zhi Chen, Pei Chu, et~al.
\newblock Internlm2 technical report.
\newblock \emph{arXiv preprint arXiv:2403.17297}, 2024.

\bibitem[Chen et~al.(2024{\natexlab{a}})Chen, Li, Dong, Zhang, Zang, Chen, Duan, Wang, Qiao, Lin, et~al.]{MMStar}
Lin Chen, Jinsong Li, Xiaoyi Dong, Pan Zhang, Yuhang Zang, Zehui Chen, Haodong Duan, Jiaqi Wang, Yu Qiao, Dahua Lin, et~al.
\newblock Are we on the right way for evaluating large vision-language models?
\newblock \emph{arXiv preprint arXiv:2403.20330}, 2024{\natexlab{a}}.

\bibitem[Chen et~al.(2024{\natexlab{b}})Chen, Deng, Yuan, Ji, and Gu]{chen2024self}
Zixiang Chen, Yihe Deng, Huizhuo Yuan, Kaixuan Ji, and Quanquan Gu.
\newblock Self-play fine-tuning converts weak language models to strong language models.
\newblock \emph{arXiv preprint arXiv:2401.01335}, 2024{\natexlab{b}}.

\bibitem[Chen et~al.(2024{\natexlab{c}})Chen, Wang, Cao, Liu, Gao, Cui, Zhu, Ye, Tian, Liu, et~al.]{InternVL2.5}
Zhe Chen, Weiyun Wang, Yue Cao, Yangzhou Liu, Zhangwei Gao, Erfei Cui, Jinguo Zhu, Shenglong Ye, Hao Tian, Zhaoyang Liu, et~al.
\newblock Expanding performance boundaries of open-source multimodal models with model, data, and test-time scaling.
\newblock \emph{arXiv preprint arXiv:2412.05271}, 2024{\natexlab{c}}.

\bibitem[Dong et~al.(2024)Dong, Liu, Sun, Yang, Hu, Rao, and Liu]{insightv}
Yuhao Dong, Zuyan Liu, Hai-Long Sun, Jingkang Yang, Winston Hu, Yongming Rao, and Ziwei Liu.
\newblock Insight-v: Exploring long-chain visual reasoning with multimodal large language models.
\newblock \emph{arXiv preprint arXiv:2411.14432}, 2024.

\bibitem[Duan et~al.(2024)Duan, Yang, Qiao, Fang, Chen, Liu, Dong, Zang, Zhang, Wang, et~al.]{vlmevalkit}
Haodong Duan, Junming Yang, Yuxuan Qiao, Xinyu Fang, Lin Chen, Yuan Liu, Xiaoyi Dong, Yuhang Zang, Pan Zhang, Jiaqi Wang, et~al.
\newblock Vlmevalkit: An open-source toolkit for evaluating large multi-modality models.
\newblock In \emph{Proceedings of the 32nd ACM international conference on multimedia}, pages 11198--11201, 2024.

\bibitem[Dubey et~al.(2024)Dubey, Jauhri, Pandey, Kadian, Al-Dahle, Letman, Mathur, Schelten, Yang, Fan, et~al.]{llama3}
Abhimanyu Dubey, Abhinav Jauhri, Abhinav Pandey, Abhishek Kadian, Ahmad Al-Dahle, Aiesha Letman, Akhil Mathur, Alan Schelten, Amy Yang, Angela Fan, et~al.
\newblock The llama 3 herd of models.
\newblock \emph{arXiv preprint arXiv:2407.21783}, 2024.

\bibitem[Dubois et~al.(2024)Dubois, Galambosi, Liang, and Hashimoto]{wr1alpacaeval2}
Yann Dubois, Bal{\'a}zs Galambosi, Percy Liang, and Tatsunori~B Hashimoto.
\newblock Length-controlled alpacaeval: A simple way to debias automatic evaluators.
\newblock \emph{arXiv preprint arXiv:2404.04475}, 2024.

\bibitem[Fu et~al.(2024)Fu, Hu, Li, Feng, Wang, Lin, Roth, Smith, Ma, and Krishna]{fu2024blink}
Xingyu Fu, Yushi Hu, Bangzheng Li, Yu Feng, Haoyu Wang, Xudong Lin, Dan Roth, Noah~A Smith, Wei-Chiu Ma, and Ranjay Krishna.
\newblock Blink: Multimodal large language models can see but not perceive.
\newblock In \emph{European Conference on Computer Vision}, pages 148--166. Springer, 2024.

\bibitem[Guan et~al.(2024)Guan, Liu, Wu, Xian, Li, Liu, Wang, Chen, Huang, Yacoob, et~al.]{guan2024hallusionbench}
Tianrui Guan, Fuxiao Liu, Xiyang Wu, Ruiqi Xian, Zongxia Li, Xiaoyu Liu, Xijun Wang, Lichang Chen, Furong Huang, Yaser Yacoob, et~al.
\newblock Hallusionbench: an advanced diagnostic suite for entangled language hallucination and visual illusion in large vision-language models.
\newblock In \emph{Proceedings of the IEEE/CVF Conference on Computer Vision and Pattern Recognition}, pages 14375--14385, 2024.

\bibitem[Jiang et~al.(2024)Jiang, He, Zeng, Wei, Ku, Liu, and Chen]{Jiang2024MANTISIM}
Dongfu Jiang, Xuan He, Huaye Zeng, Cong Wei, Max~W.F. Ku, Qian Liu, and Wenhu Chen.
\newblock Mantis: Interleaved multi-image instruction tuning.
\newblock \emph{Transactions on Machine Learning Research}, 2024, 2024.

\bibitem[Khan et~al.(2025)Khan, Leem, See, Wong, Zhang, and Fang]{LvlmMedicalSurvey}
Wasif Khan, Seowung Leem, Kyle~B See, Joshua~K Wong, Shaoting Zhang, and Ruogu Fang.
\newblock A comprehensive survey of foundation models in medicine.
\newblock \emph{IEEE Reviews in Biomedical Engineering}, 2025.

\bibitem[Kim et~al.(2023)Kim, Park, Kim, Lee, Song, Kim, Kim, Kim, Lee, Kim, et~al.]{kim2023solar}
Dahyun Kim, Chanjun Park, Sanghoon Kim, Wonsung Lee, Wonho Song, Yunsu Kim, Hyeonwoo Kim, Yungi Kim, Hyeonju Lee, Jihoo Kim, et~al.
\newblock Solar 10.7 b: Scaling large language models with simple yet effective depth up-scaling.
\newblock \emph{arXiv preprint arXiv:2312.15166}, 2023.

\bibitem[Li et~al.(2024{\natexlab{a}})Li, Zhang, Guo, Zhang, Li, Zhang, Zhang, Zhang, Li, Liu, et~al.]{Llava-OneVision}
Bo Li, Yuanhan Zhang, Dong Guo, Renrui Zhang, Feng Li, Hao Zhang, Kaichen Zhang, Peiyuan Zhang, Yanwei Li, Ziwei Liu, et~al.
\newblock Llava-onevision: Easy visual task transfer.
\newblock \emph{arXiv preprint arXiv:2408.03326}, 2024{\natexlab{a}}.

\bibitem[Li et~al.(2024{\natexlab{b}})Li, Wei, Xie, Yang, Song, Wang, An, Liu, Li, Lin, et~al.]{VLRewardBench}
Lei Li, Yuancheng Wei, Zhihui Xie, Xuqing Yang, Yifan Song, Peiyi Wang, Chenxin An, Tianyu Liu, Sujian Li, Bill~Yuchen Lin, et~al.
\newblock Vlrewardbench: A challenging benchmark for vision-language generative reward models.
\newblock \emph{arXiv preprint arXiv:2411.17451}, 2024{\natexlab{b}}.

\bibitem[Li et~al.(2024{\natexlab{c}})Li, Xie, Li, Chen, Wang, Chen, Yang, Wang, Kong, and Liu]{vlfeedback}
Lei Li, Zhihui Xie, Mukai Li, Shunian Chen, Peiyi Wang, Liang Chen, Yazheng Yang, Benyou Wang, Lingpeng Kong, and Qi Liu.
\newblock Vlfeedback: A large-scale ai feedback dataset for large vision-language models alignment.
\newblock \emph{arXiv preprint arXiv:2410.09421}, 2024{\natexlab{c}}.

\bibitem[Li et~al.(2025)Li, Wu, Du, Nghiem, and Shi]{lvlmsurvey}
Zongxia Li, Xiyang Wu, Hongyang Du, Huy Nghiem, and Guangyao Shi.
\newblock Benchmark evaluations, applications, and challenges of large vision language models: A survey.
\newblock \emph{arXiv preprint arXiv:2501.02189}, 2025.

\bibitem[Lightman et~al.(2023)Lightman, Kosaraju, Burda, Edwards, Baker, Lee, Leike, Schulman, Sutskever, and Cobbe]{prm800k}
Hunter Lightman, Vineet Kosaraju, Yuri Burda, Harrison Edwards, Bowen Baker, Teddy Lee, Jan Leike, John Schulman, Ilya Sutskever, and Karl Cobbe.
\newblock Let's verify step by step.
\newblock In \emph{The Twelfth International Conference on Learning Representations}, 2023.

\bibitem[Lin et~al.(2024)Lin, Gou, Liang, Luo, Liu, and Yang]{CriticBench}
Zicheng Lin, Zhibin Gou, Tian Liang, Ruilin Luo, Haowei Liu, and Yujiu Yang.
\newblock Criticbench: Benchmarking llms for critique-correct reasoning.
\newblock \emph{arXiv preprint arXiv:2402.14809}, 2024.

\bibitem[Liu et~al.(2024{\natexlab{a}})Liu, Zeng, Liu, Yan, He, Wang, Yan, Liu, and Zhou]{liu2024skywork}
Chris~Yuhao Liu, Liang Zeng, Jiacai Liu, Rui Yan, Jujie He, Chaojie Wang, Shuicheng Yan, Yang Liu, and Yahui Zhou.
\newblock Skywork-reward: Bag of tricks for reward modeling in llms.
\newblock \emph{arXiv preprint arXiv:2410.18451}, 2024{\natexlab{a}}.

\bibitem[Liu et~al.(2024{\natexlab{b}})Liu, Xue, Chen, Chen, Zhao, Wang, Hou, Li, and Peng]{LvlmHallucinationSurvey}
Hanchao Liu, Wenyuan Xue, Yifei Chen, Dapeng Chen, Xiutian Zhao, Ke Wang, Liping Hou, Rongjun Li, and Wei Peng.
\newblock A survey on hallucination in large vision-language models.
\newblock \emph{arXiv preprint arXiv:2402.00253}, 2024{\natexlab{b}}.

\bibitem[Liu et~al.(2024{\natexlab{c}})Liu, Yao, Min, Cao, Hou, and Li]{RMBench}
Yantao Liu, Zijun Yao, Rui Min, Yixin Cao, Lei Hou, and Juanzi Li.
\newblock Rm-bench: Benchmarking reward models of language models with subtlety and style.
\newblock \emph{arXiv preprint arXiv:2410.16184}, 2024{\natexlab{c}}.

\bibitem[Lou et~al.(2024)Lou, Yan, Shen, Yan, Xie, and Zhang]{lou2024uncertainty}
Xingzhou Lou, Dong Yan, Wei Shen, Yuzi Yan, Jian Xie, and Junge Zhang.
\newblock Uncertainty-aware reward model: Teaching reward models to know what is unknown.
\newblock \emph{arXiv preprint arXiv:2410.00847}, 2024.

\bibitem[Lu et~al.(2023)Lu, Bansal, Xia, Liu, Li, Hajishirzi, Cheng, Chang, Galley, and Gao]{lu2023mathvista}
Pan Lu, Hritik Bansal, Tony Xia, Jiacheng Liu, Chunyuan Li, Hannaneh Hajishirzi, Hao Cheng, Kai-Wei Chang, Michel Galley, and Jianfeng Gao.
\newblock Mathvista: Evaluating mathematical reasoning of foundation models in visual contexts.
\newblock \emph{arXiv preprint arXiv:2310.02255}, 2023.

\bibitem[Lu et~al.(2024)Lu, Li, Chen, Xu, Luo, Zhang, and Ye]{Ovis2}
Shiyin Lu, Yang Li, Qing-Guo Chen, Zhao Xu, Weihua Luo, Kaifu Zhang, and Han-Jia Ye.
\newblock Ovis: Structural embedding alignment for multimodal large language model.
\newblock \emph{arXiv preprint arXiv:2405.20797}, 2024.

\bibitem[Ma et~al.(2023)Ma, Zhou, Liu, Yuan, Liu, You, and Yang]{ma2023let}
Qianli Ma, Haotian Zhou, Tingkai Liu, Jianbo Yuan, Pengfei Liu, Yang You, and Hongxia Yang.
\newblock Let's reward step by step: Step-level reward model as the navigators for reasoning.
\newblock \emph{arXiv preprint arXiv:2310.10080}, 2023.

\bibitem[Meta(2024)]{llama3onlyL}
AI Meta.
\newblock Introducing meta llama 3: The most capable openly available llm to date.
\newblock \emph{Meta AI}, 2024.

\bibitem[OpenBMB(2025)]{MiniCPM}
OpenBMB.
\newblock Minicpm o: A gpt-4o-level mllm for vision, speech, and multimodal live streaming on your phone, 2025.
\newblock Accessed: 2025-02-25.

\bibitem[Ouyang et~al.(2022)Ouyang, Wu, Jiang, Almeida, Wainwright, Mishkin, Zhang, Agarwal, Slama, Ray, et~al.]{rlhfRM2}
Long Ouyang, Jeffrey Wu, Xu Jiang, Diogo Almeida, Carroll Wainwright, Pamela Mishkin, Chong Zhang, Sandhini Agarwal, Katarina Slama, Alex Ray, et~al.
\newblock Training language models to follow instructions with human feedback.
\newblock \emph{Advances in neural information processing systems}, 35:\penalty0 27730--27744, 2022.

\bibitem[Rafailov et~al.(2023)Rafailov, Sharma, Mitchell, Manning, Ermon, and Finn]{dpo}
Rafael Rafailov, Archit Sharma, Eric Mitchell, Christopher~D Manning, Stefano Ermon, and Chelsea Finn.
\newblock Direct preference optimization: Your language model is secretly a reward model.
\newblock \emph{Advances in Neural Information Processing Systems}, 36:\penalty0 53728--53741, 2023.

\bibitem[Ruan et~al.(2024{\natexlab{a}})Ruan, Yang, Lin, Feng, Xiong, Tang, and Li]{ftiibench}
Jiacheng Ruan, Yebin Yang, Zehao Lin, Yuchen Feng, Feiyu Xiong, Zeyun Tang, and Zhiyu Li.
\newblock Ftii-bench: A comprehensive multimodal benchmark for flow text with image insertion.
\newblock \emph{arXiv preprint arXiv:2410.12564}, 2024{\natexlab{a}}.

\bibitem[Ruan et~al.(2024{\natexlab{b}})Ruan, Yuan, Lin, Liao, Li, Xiong, Liu, and Fu]{camobj}
Jiacheng Ruan, Wenzhen Yuan, Zehao Lin, Ning Liao, Zhiyu Li, Feiyu Xiong, Ting Liu, and Yuzhuo Fu.
\newblock Mm-camobj: A comprehensive multimodal dataset for camouflaged object scenarios.
\newblock \emph{arXiv preprint arXiv:2409.16084}, 2024{\natexlab{b}}.

\bibitem[Shahgir et~al.(2024)Shahgir, Sayeed, Bhattacharjee, Ahmad, Dong, and Shahriyar]{shahgir2024illusionvqa}
Haz~Sameen Shahgir, Khondker~Salman Sayeed, Abhik Bhattacharjee, Wasi~Uddin Ahmad, Yue Dong, and Rifat Shahriyar.
\newblock Illusionvqa: A challenging optical illusion dataset for vision language models.
\newblock \emph{arXiv preprint arXiv:2403.15952}, 2024.

\bibitem[Song et~al.(2025)Song, Su, Qu, Zhou, and Cheng]{PRMBench}
Mingyang Song, Zhaochen Su, Xiaoye Qu, Jiawei Zhou, and Yu Cheng.
\newblock Prmbench: A fine-grained and challenging benchmark for process-level reward models.
\newblock \emph{arXiv preprint arXiv:2501.03124}, 2025.

\bibitem[Team et~al.(2023)Team, Anil, Borgeaud, Alayrac, Yu, Soricut, Schalkwyk, Dai, Hauth, Millican, et~al.]{gemini}
Gemini Team, Rohan Anil, Sebastian Borgeaud, Jean-Baptiste Alayrac, Jiahui Yu, Radu Soricut, Johan Schalkwyk, Andrew~M Dai, Anja Hauth, Katie Millican, et~al.
\newblock Gemini: a family of highly capable multimodal models.
\newblock \emph{arXiv preprint arXiv:2312.11805}, 2023.

\bibitem[Team(2024)]{qvq-72b-preview}
Qwen Team.
\newblock Qvq: To see the world with wisdom, 2024.

\bibitem[Touvron et~al.(2023)Touvron, Martin, Stone, Albert, Almahairi, Babaei, Bashlykov, Batra, Bhargava, Bhosale, et~al.]{llama2}
Hugo Touvron, Louis Martin, Kevin Stone, Peter Albert, Amjad Almahairi, Yasmine Babaei, Nikolay Bashlykov, Soumya Batra, Prajjwal Bhargava, Shruti Bhosale, et~al.
\newblock Llama 2: Open foundation and fine-tuned chat models.
\newblock \emph{arXiv preprint arXiv:2307.09288}, 2023.

\bibitem[Wang et~al.(2024{\natexlab{a}})Wang, Zhang, Du, Zhang, and Chu]{DataSelectionSurvey}
Jiahao Wang, Bolin Zhang, Qianlong Du, Jiajun Zhang, and Dianhui Chu.
\newblock A survey on data selection for llm instruction tuning.
\newblock \emph{arXiv preprint arXiv:2402.05123}, 2024{\natexlab{a}}.

\bibitem[Wang et~al.(2025)Wang, Pan, Shi, Lu, Ren, Zhou, Zhan, and Li]{wang2025measuring}
Ke Wang, Junting Pan, Weikang Shi, Zimu Lu, Houxing Ren, Aojun Zhou, Mingjie Zhan, and Hongsheng Li.
\newblock Measuring multimodal mathematical reasoning with math-vision dataset.
\newblock \emph{Advances in Neural Information Processing Systems}, 37:\penalty0 95095--95169, 2025.

\bibitem[Wang et~al.(2023)Wang, Li, Chen, Cai, Zhu, Lin, Cao, Liu, Liu, and Sui]{sampleorder}
Peiyi Wang, Lei Li, Liang Chen, Zefan Cai, Dawei Zhu, Binghuai Lin, Yunbo Cao, Qi Liu, Tianyu Liu, and Zhifang Sui.
\newblock Large language models are not fair evaluators.
\newblock \emph{arXiv preprint arXiv:2305.17926}, 2023.

\bibitem[Wang et~al.(2024{\natexlab{b}})Wang, Bai, Tan, Wang, Fan, Bai, Chen, Liu, Wang, Ge, et~al.]{Qwen2-VL}
Peng Wang, Shuai Bai, Sinan Tan, Shijie Wang, Zhihao Fan, Jinze Bai, Keqin Chen, Xuejing Liu, Jialin Wang, Wenbin Ge, et~al.
\newblock Qwen2-vl: Enhancing vision-language model's perception of the world at any resolution.
\newblock \emph{arXiv preprint arXiv:2409.12191}, 2024{\natexlab{b}}.

\bibitem[Wang et~al.(2024{\natexlab{c}})Wang, Kulikov, Golovneva, Yu, Yuan, Dwivedi-Yu, Pang, Fazel-Zarandi, Weston, and Li]{wang2024self}
Tianlu Wang, Ilia Kulikov, Olga Golovneva, Ping Yu, Weizhe Yuan, Jane Dwivedi-Yu, Richard~Yuanzhe Pang, Maryam Fazel-Zarandi, Jason Weston, and Xian Li.
\newblock Self-taught evaluators.
\newblock \emph{arXiv preprint arXiv:2408.02666}, 2024{\natexlab{c}}.

\bibitem[Wang et~al.(2024{\natexlab{d}})Wang, Bukharin, Delalleau, Egert, Shen, Zeng, Kuchaiev, and Dong]{helpsteer2}
Zhilin Wang, Alexander Bukharin, Olivier Delalleau, Daniel Egert, Gerald Shen, Jiaqi Zeng, Oleksii Kuchaiev, and Yi Dong.
\newblock Helpsteer2-preference: Complementing ratings with preferences.
\newblock \emph{arXiv preprint arXiv:2410.01257}, 2024{\natexlab{d}}.

\bibitem[Wu et~al.(2025)Wu, Feng, Zhang, Jin, Che, Wen, and Tao]{vlmcts}
Jinyang Wu, Mingkuan Feng, Shuai Zhang, Ruihan Jin, Feihu Che, Zengqi Wen, and Jianhua Tao.
\newblock Boosting multimodal reasoning with mcts-automated structured thinking.
\newblock \emph{arXiv preprint arXiv:2502.02339}, 2025.

\bibitem[Xiyao et~al.(2024)Xiyao, Zhengyuan, Linjie, Hongjin, Yuancheng, Lin, Kevin, Furong, and Lijuan]{visvm}
Wang Xiyao, Yang Zhengyuan, Li Linjie, Lu Hongjin, Xu Yuancheng, Lin Chung-Ching Lin, Lin Kevin, Huang Furong, and Wang Lijuan.
\newblock Scaling inference-time search with vision value model for improved visual comprehension.
\newblock \emph{arXiv preprint arXiv:2412.03704}, 2024.

\bibitem[Yang et~al.(2024{\natexlab{a}})Yang, Yang, Hui, Zheng, Yu, Zhou, Li, Li, Liu, Huang, et~al.]{qwen2}
An Yang, Baosong Yang, Binyuan Hui, Bo Zheng, Bowen Yu, Chang Zhou, Chengpeng Li, Chengyuan Li, Dayiheng Liu, Fei Huang, et~al.
\newblock Qwen2 technical report.
\newblock \emph{arXiv preprint arXiv:2407.10671}, 2024{\natexlab{a}}.

\bibitem[Yang et~al.(2024{\natexlab{b}})Yang, Yang, Zhang, Hui, Zheng, Yu, Li, Liu, Huang, Wei, et~al.]{qwen25}
An Yang, Baosong Yang, Beichen Zhang, Binyuan Hui, Bo Zheng, Bowen Yu, Chengyuan Li, Dayiheng Liu, Fei Huang, Haoran Wei, et~al.
\newblock Qwen2. 5 technical report.
\newblock \emph{arXiv preprint arXiv:2412.15115}, 2024{\natexlab{b}}.

\bibitem[Yao et~al.(2024)Yao, Huang, Wu, Zhang, Wang, Liu, Wang, Song, Feng, Shen, et~al.]{mulberry}
Huanjin Yao, Jiaxing Huang, Wenhao Wu, Jingyi Zhang, Yibo Wang, Shunyu Liu, Yingjie Wang, Yuxin Song, Haocheng Feng, Li Shen, et~al.
\newblock Mulberry: Empowering mllm with o1-like reasoning and reflection via collective monte carlo tree search.
\newblock \emph{arXiv preprint arXiv:2412.18319}, 2024.

\bibitem[Yasunaga et~al.(2025)Yasunaga, Zettlemoyer, and Ghazvininejad]{mmrewardbench}
Michihiro Yasunaga, Luke Zettlemoyer, and Marjan Ghazvininejad.
\newblock Multimodal rewardbench: Holistic evaluation of reward models for vision language models.
\newblock \emph{arXiv preprint arXiv:2502.14191}, 2025.

\bibitem[Yu et~al.(2024)Yu, Zhang, Yao, Dang, Chen, Lu, Cui, He, Liu, Chua, et~al.]{rlaif}
Tianyu Yu, Haoye Zhang, Yuan Yao, Yunkai Dang, Da Chen, Xiaoman Lu, Ganqu Cui, Taiwen He, Zhiyuan Liu, Tat-Seng Chua, et~al.
\newblock Rlaif-v: Aligning mllms through open-source ai feedback for super gpt-4v trustworthiness.
\newblock \emph{arXiv preprint arXiv:2405.17220}, 2024.

\bibitem[Zeng et~al.(2023)Zeng, Chen, Liu, Jiang, and Jia]{MRGSM8k}
Zhongshen Zeng, Pengguang Chen, Shu Liu, Haiyun Jiang, and Jiaya Jia.
\newblock Mr-gsm8k: A meta-reasoning benchmark for large language model evaluation.
\newblock \emph{arXiv preprint arXiv:2312.17080}, 2023.

\bibitem[Zeng et~al.(2024)Zeng, Liu, Wan, Li, Chen, Dai, Yao, Xu, Qi, Zhao, et~al.]{mrben}
Zhongshen Zeng, Yinhong Liu, Yingjia Wan, Jingyao Li, Pengguang Chen, Jianbo Dai, Yuxuan Yao, Rongwu Xu, Zehan Qi, Wanru Zhao, et~al.
\newblock Mr-ben: A meta-reasoning benchmark for evaluating system-2 thinking in llms.
\newblock \emph{arXiv preprint arXiv:2406.13975}, 2024.

\bibitem[Zhan et~al.(2025)Zhan, Xiong, and Yuan]{skyeyegpt}
Yang Zhan, Zhitong Xiong, and Yuan Yuan.
\newblock Skyeyegpt: Unifying remote sensing vision-language tasks via instruction tuning with large language model.
\newblock \emph{ISPRS Journal of Photogrammetry and Remote Sensing}, 221:\penalty0 64--77, 2025.

\bibitem[Zhang et~al.(2024{\natexlab{a}})Zhang, Lei, Li, Wang, Liu, Yang, Li, Wang, Yang, Wu, et~al.]{criticv}
Di Zhang, Jingdi Lei, Junxian Li, Xunzhi Wang, Yujie Liu, Zonglin Yang, Jiatong Li, Weida Wang, Suorong Yang, Jianbo Wu, et~al.
\newblock Critic-v: Vlm critics help catch vlm errors in multimodal reasoning.
\newblock \emph{arXiv preprint arXiv:2411.18203}, 2024{\natexlab{a}}.

\bibitem[Zhang et~al.(2024{\natexlab{b}})Zhang, Jiang, Zhang, Lin, Guo, Qiu, Zhou, Lu, Chang, Qiao, et~al.]{zhang2024mathverse}
Renrui Zhang, Dongzhi Jiang, Yichi Zhang, Haokun Lin, Ziyu Guo, Pengshuo Qiu, Aojun Zhou, Pan Lu, Kai-Wei Chang, Yu Qiao, et~al.
\newblock Mathverse: Does your multi-modal llm truly see the diagrams in visual math problems?
\newblock In \emph{European Conference on Computer Vision}, pages 169--186. Springer, 2024{\natexlab{b}}.

\bibitem[Zhang et~al.(2025)Zhang, Zheng, Wu, Zhang, Lin, Yu, Liu, Zhou, and Lin]{zhang2025lessons}
Zhenru Zhang, Chujie Zheng, Yangzhen Wu, Beichen Zhang, Runji Lin, Bowen Yu, Dayiheng Liu, Jingren Zhou, and Junyang Lin.
\newblock The lessons of developing process reward models in mathematical reasoning.
\newblock \emph{arXiv preprint arXiv:2501.07301}, 2025.

\bibitem[Zheng et~al.(2024)Zheng, Zhang, Zhang, Lin, Lu, Yu, Liu, Zhou, and Lin]{qwenprocessbench}
Chujie Zheng, Zhenru Zhang, Beichen Zhang, Runji Lin, Keming Lu, Bowen Yu, Dayiheng Liu, Jingren Zhou, and Junyang Lin.
\newblock Processbench: Identifying process errors in mathematical reasoning.
\newblock \emph{arXiv preprint arXiv:2412.06559}, 2024.

\bibitem[Zheng et~al.(2023)Zheng, Chiang, Sheng, Zhuang, Wu, Zhuang, Lin, Li, Li, Xing, et~al.]{wr2mtbench}
Lianmin Zheng, Wei-Lin Chiang, Ying Sheng, Siyuan Zhuang, Zhanghao Wu, Yonghao Zhuang, Zi Lin, Zhuohan Li, Dacheng Li, Eric Xing, et~al.
\newblock Judging llm-as-a-judge with mt-bench and chatbot arena.
\newblock \emph{Advances in Neural Information Processing Systems}, 36:\penalty0 46595--46623, 2023.

\bibitem[Zhou et~al.(2024{\natexlab{a}})Zhou, Liu, Yurtsever, Zagar, Zimmer, Cao, and Knoll]{LvlmAutoDriveSurvey}
Xingcheng Zhou, Mingyu Liu, Ekim Yurtsever, Bare~Luka Zagar, Walter Zimmer, Hu Cao, and Alois~C Knoll.
\newblock Vision language models in autonomous driving: A survey and outlook.
\newblock \emph{IEEE Transactions on Intelligent Vehicles}, 2024{\natexlab{a}}.

\bibitem[Zhou et~al.(2024{\natexlab{b}})Zhou, Liu, Ning, Liu, Wang, Wong, Huang, Wang, and Huang]{MathCheck_GSM}
Zihao Zhou, Shudong Liu, Maizhen Ning, Wei Liu, Jindong Wang, Derek~F Wong, Xiaowei Huang, Qiufeng Wang, and Kaizhu Huang.
\newblock Is your model really a good math reasoner? evaluating mathematical reasoning with checklist.
\newblock \emph{arXiv preprint arXiv:2407.08733}, 2024{\natexlab{b}}.

\bibitem[Zhu et~al.(2024)Zhu, Qu, Dong, Ruan, Tong, He, and Cheng]{llamamoe}
Tong Zhu, Xiaoye Qu, Daize Dong, Jiacheng Ruan, Jingqi Tong, Conghui He, and Yu Cheng.
\newblock Llama-moe: Building mixture-of-experts from llama with continual pre-training.
\newblock In \emph{Proceedings of the 2024 Conference on Empirical Methods in Natural Language Processing}, pages 15913--15923, 2024.

\bibitem[Zou et~al.(2024)Zou, Guo, Yang, Zhang, Hu, and Zhang]{zou2024dynamath}
Chengke Zou, Xingang Guo, Rui Yang, Junyu Zhang, Bin Hu, and Huan Zhang.
\newblock Dynamath: A dynamic visual benchmark for evaluating mathematical reasoning robustness of vision language models.
\newblock \emph{arXiv preprint arXiv:2411.00836}, 2024.

\end{thebibliography}
